\documentclass[letterpaper]{article} 
\usepackage{aaai25}  
\usepackage{times}  
\usepackage{helvet}  
\usepackage{courier}  
\usepackage[hyphens]{url}  
\usepackage{graphicx} 
\usepackage{natbib}  
\usepackage{caption} 
\usepackage{algorithm}
\usepackage{algorithmic}

\usepackage{svg}

\usepackage{amsmath}
\usepackage{comment}
\usepackage{amsfonts}
\usepackage{multirow}
\usepackage{rotating}
\usepackage{booktabs}
\usepackage{subfigure}

\usepackage{newfloat}
\usepackage{listings}
\DeclareCaptionStyle{ruled}{labelfont=normalfont,labelsep=colon,strut=off} 
\floatstyle{ruled}
\newfloat{listing}{tb}{lst}{}
\floatname{listing}{Listing}
\title{Is LLMs Hallucination Usable? \\ LLM-based Negative Reasoning for Fake News Detection}
\author{
    Chaowei Zhang\textsuperscript{\rm 1}, 
    Zongling Feng\textsuperscript{\rm 1}, 
    Zewei Zhang\textsuperscript{\rm 2},  
    Jipeng Qiang\textsuperscript{\rm 1},  
    Guandong Xu\textsuperscript{\rm 3}, 
    Yun Li \textsuperscript{\rm 1}\thanks{Yun Li is the corresponding Author.}
}
\affiliations{

    \textsuperscript{\rm 1}Yangzhou University\\
    \textsuperscript{\rm 2}Auburn University\\
    \textsuperscript{\rm 3}The Education University of Hong Kong

    \{cwzhang,jpqiang,liyun\}@yzu.edu.cn,\\
    mz120220995@stu.yzu.edu.cn,
    zzw475622@gmail.com,
    gdxu@eduhk.hk,
    
}

\usepackage{bibentry}

\begin{document}

\maketitle

\begin{abstract}
The questionable responses caused by knowledge hallucination may lead to LLMs' unstable ability in decision-making. However, it has never been investigated whether the LLMs' hallucination is possibly usable to generate negative reasoning for facilitating the detection of fake news. This study proposes a novel supervised self-reinforced reasoning rectification approach - SR$^3$ that yields both common reasonable reasoning and wrong understandings (negative reasoning) for news via LLMs reflection for semantic consistency learning. Upon that, we construct a negative reasoning-based news learning model called - \emph{NRFE}, which leverages positive or negative news-reasoning pairs for learning the semantic consistency between them. To avoid the impact of label-implicated reasoning, we deploy a student model - \emph{NRFE-D} that only takes news content as input to inspect the performance of our method by distilling the knowledge from \emph{NRFE}. The experimental results verified on three popular fake news datasets demonstrate the superiority of our method compared with three kinds of baselines including prompting on LLMs, fine-tuning on pre-trained SLMs, and other representative fake news detection methods. 
\end{abstract}

%
\section{Introduction}
Recently, Large Language Models (LLMs), such as GPT-4o, Claude 3, and Llama 3.1 represent a significant lead forward in the field of AI, especially for NLP. The LLMs trained on vast amounts of human-generated text can deeply understand and interpret prompts as well as generate comprehensive, coherent, and contextually relevant reasoning for the prompts~\cite{yang2024harnessing,bang-etal-2023-multitask}, which allow them to be applicable to various NLP downstream tasks including fake news detection. Specifically, LLMs can provide multi-aspect aids for fake news detection, including contextual analysis, content understanding, fact-checking, source verification, reasoning generation, and others~\cite{liu2024make,pelrine-etal-2023-towards}. 
However, some limitations caused by the problem of "knowledge hallucination"~\cite{chen2023hallucination} are still unavoidable while applying LLMs to identify fake news. The up-to-date related studies are devoted to eliminating hallucinations via RAG (Retrieval-Augmented Generation)~\cite{boumber2024llms}, knowledge graph~\cite{guo2023interpretable}, or other tricks, yet it is a kind of food for thought that \textbf{is it possible to use hallucination to do something for news?} For example, can we leverage both news' correct (a.k.a., positive) reasoning and the incorrect (a.k.a., negative) reasoning generated through LLM hallucinations for adversarial semantic consistency learning?

\begin{figure}[t]
\centering
\subfigure{
\includegraphics[width=1.0\linewidth,height = 0.5\linewidth]{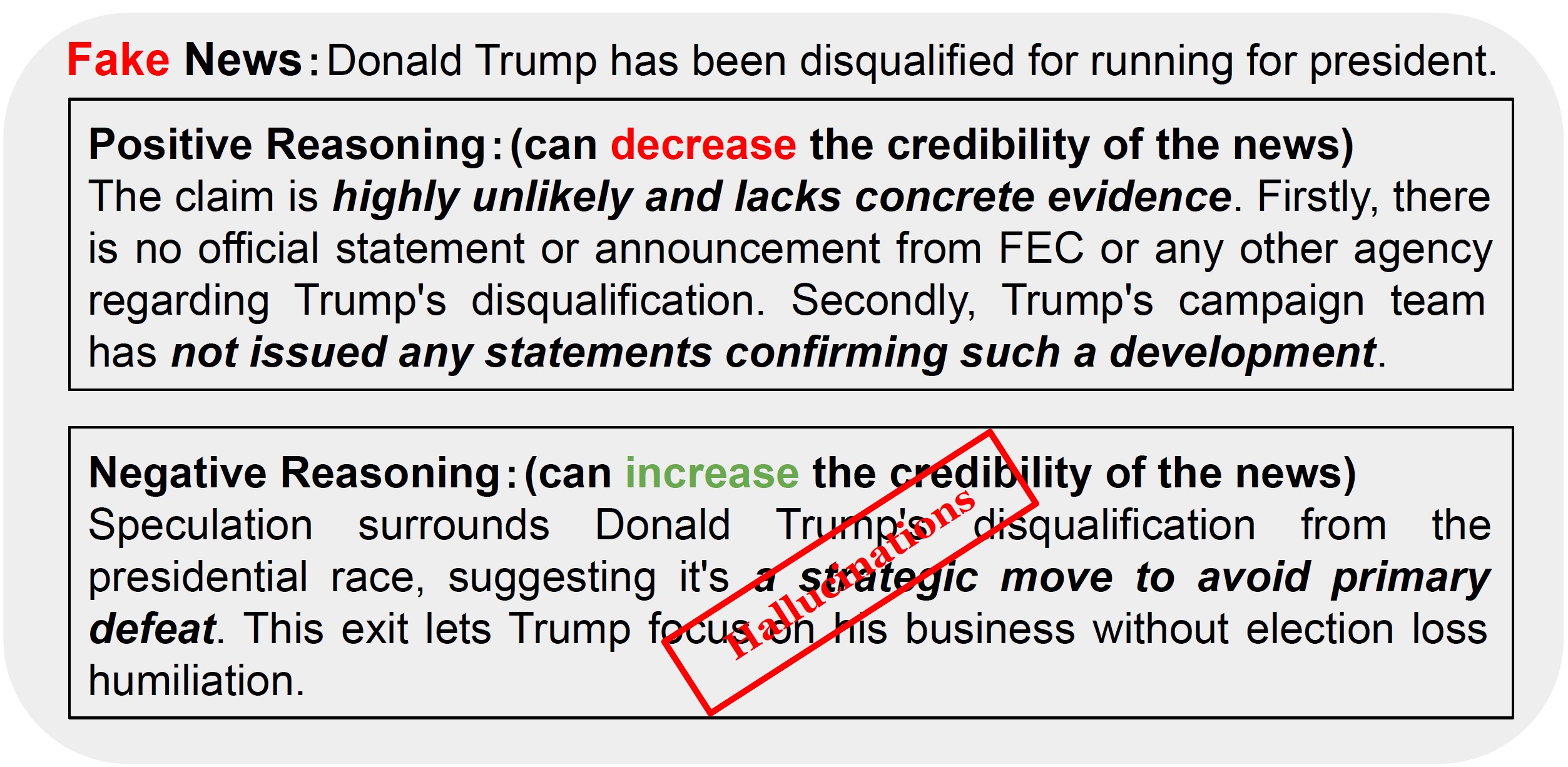}}
\subfigure{ 
\includegraphics[width=1.0\linewidth,height = 0.5\linewidth]{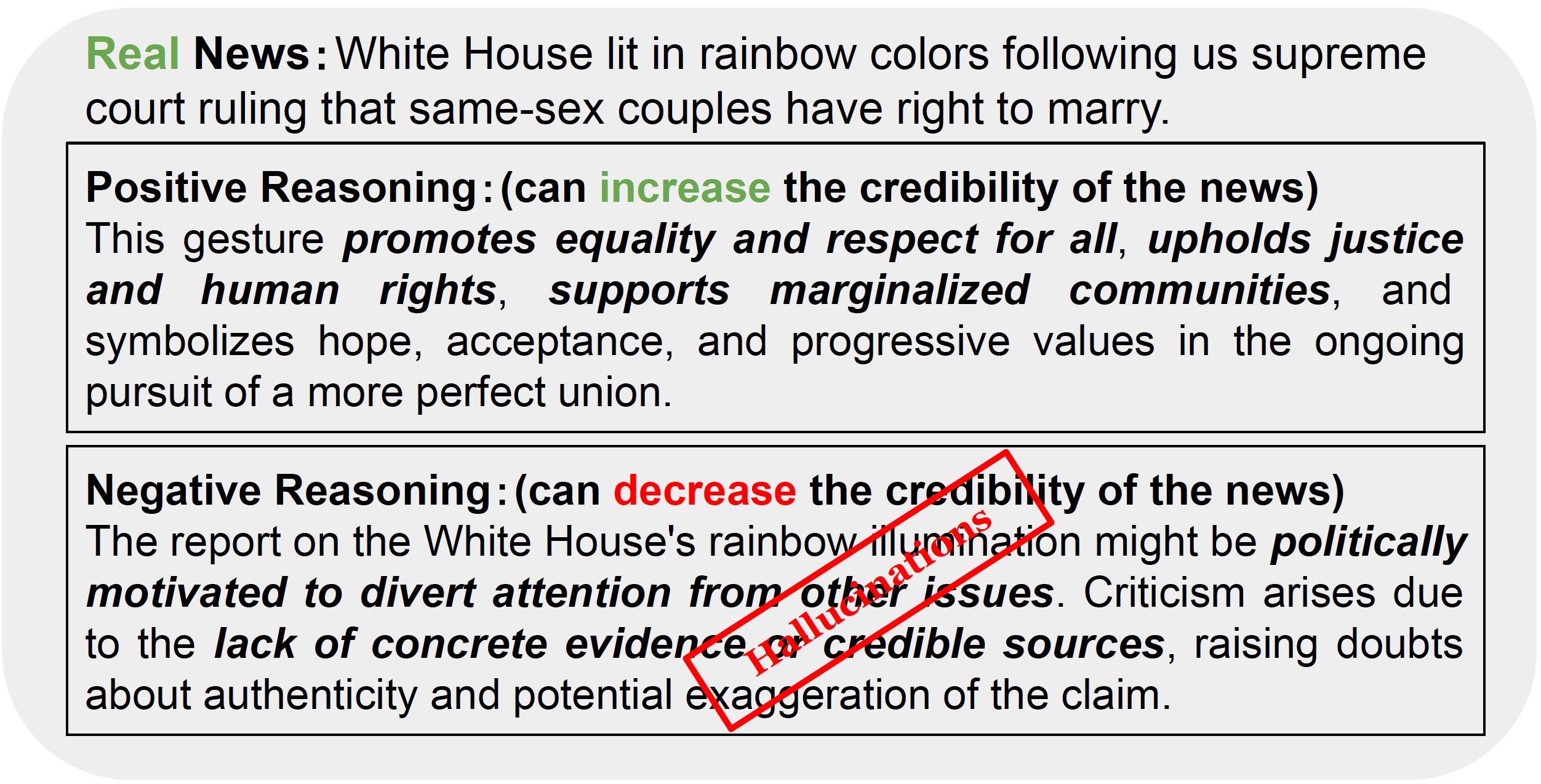}}
\caption{The examples of news' positive and negative reasoning requested via LLMs prompting. Current studies are devoted to avoiding the occurrence of negative reasoning, yet we initially propose to leverage the ability of LLM hallucinations to generate negative reasoning coupled with positive reasoning for adversarial semantic consistency learning. } 
\label{fig:examples}
\end{figure}

Generally, knowledge hallucination is a response generated by LLM, which contains "false or misleading information presented as fact"~\cite{ji2023survey}. Classical LLM reasoning usages for the tasks of fake news detection attempt to generate correct understanding (positive reasoning) as well as alleviate the impact of hallucinations via the tricks mentioned above~\cite{wang2024explainable}. However, it remains controversial that "\textbf{Are hallucinations really useless?}" As we know, the AI researchers usually take advantage of data augmentation~\cite{yin2024gamc,wu2024towards} and data perturbation~\cite{wang2023dhcf,ma2022towards} techniques to "poison" original data to strengthen the quality, and diversity of training datasets for improving the robustness and generalizability of models. Inspired by that, we raise the concern that "\textbf{Is it possible to utilize LLMs' knowledge hallucination as negative data samples?}" To verify this possibility, we initially propose to leverage LLM hallucination to generate negative reasoning for news. In this case, it would be beneficial to prevent fake news detection models from experiencing the issue of "Out-Of-Distribution"~\cite{liu2024out} by including negative reasoning of news. Fig.~\ref{fig:examples} demonstrates the examples of the generated reasoning via Llama 3. It is noteworthy that the negative reasoning is false information caused by LLM hallucinations.

To explore the ability of LLMs hallucination in generating negative reasoning for assisting fake news detection, we utilize \textit{Llama 3} to generate both positive and negative reasoning for news in a supervised manner. Specifically, qualified positive reasoning can reinforce news' polarity, yet negative reasoning is expected to decrease the confidence level and even alter the polarity of the news. To this end, we present a \textbf{S}elf-\textbf{R}einforced \textbf{R}easoning \textbf{R}ectification approach, called \textit{SR$^3$}, to force LLMs to output high-quality positive and negative reasoning by considering (1) the polarity transformation and (2) the varying confidence score of news. Based on that, we construct a negative reasoning-based news learning model - \textit{NRFE} that maintains two BERT Encoders for learning the semantic consistency between news and reasoning. To avoid the impact of supervised reasoning requests, we design a student model - \textit{NRFE-D} by distilling the knowledge from \textit{NRFE} for prediction. Overall, the contributions achieved by this study are summarized as follows:

\begin{itemize}

    \item To our best knowledge, this study originally explores the effect of LLM hallucination in generating negative reasoning aiming at fake news detection. Moreover, we propose a novel self-reinforced reasoning rectification approach (SR$^3$)
    via LLM reflection to generate qualified positive and negative reasoning for news.

    \item  We deploy a local fake news detection model complying with the design of knowledge distillation to avoid the impact of supervised reasoning requests, which contains a teacher-side module - \textit{NRFE} and a student-side module - \textit{NRFE-D}. Specifically, \textit{NRFE} involves two BERT Encoders to acquire news-reasoning pairs for semantic consistency learning, and \textit{NRFE-D} is a reasoning-free model that distills the representations of news yielded from \textit{NRFE} for prediction.

    \item The results verified on three fake news datasets demonstrate that our approach considerably outperforms three types of baselines, thereby highlighting its superiority in exploring LLMs' knowledge hallucination for generating negative reasoning.

\end{itemize}

\begin{figure}
    \centering
    \includegraphics[width=1\linewidth]{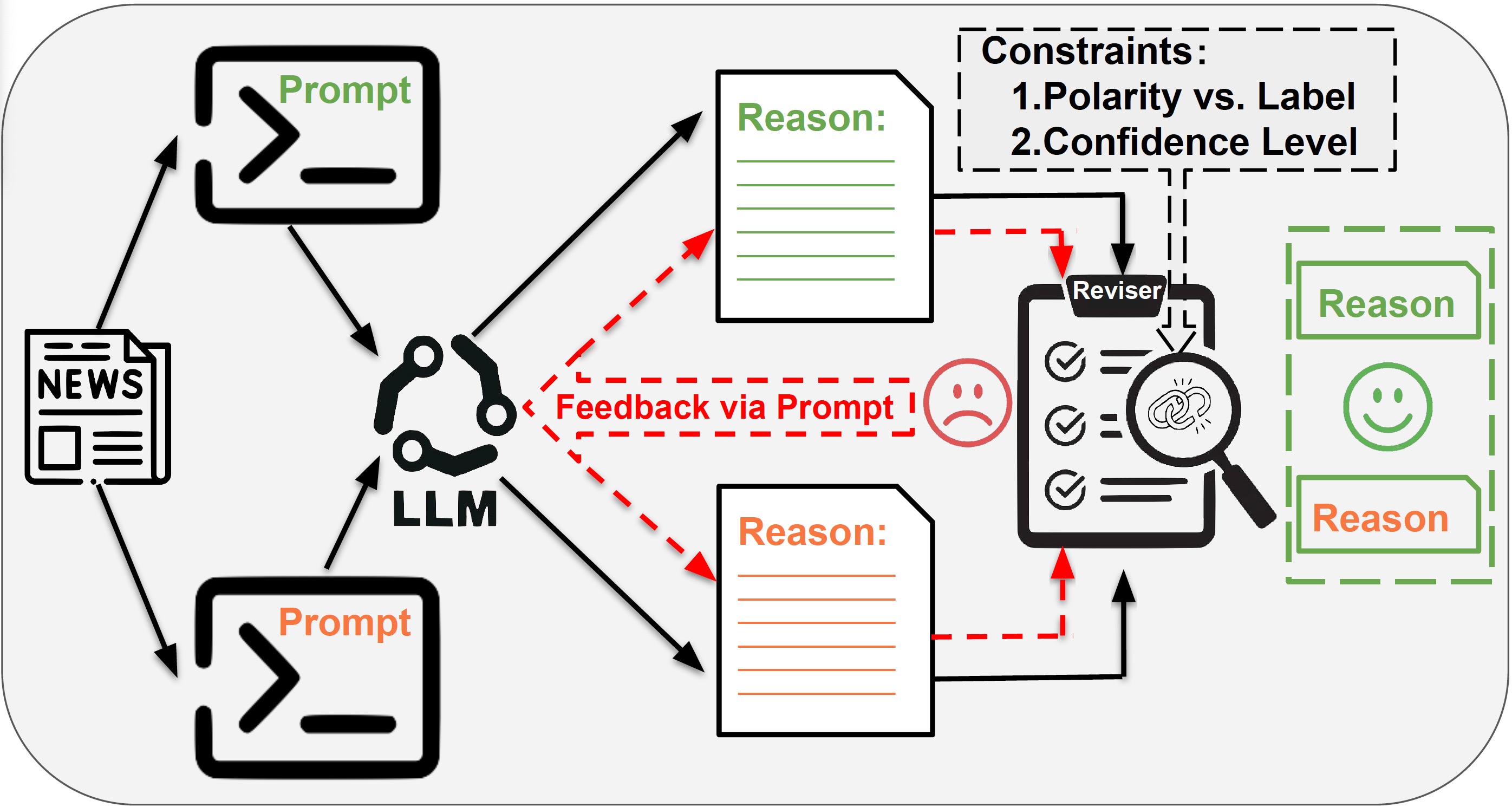}
    \caption{The demonstration of our proposed self-reinforced reasoning rectification approach - SR$^3$. Besides positive reasoning requests, SR$^3$ also utilizes the ability of LLM hallucination to iteratively generate negative reasoning until the generated negative reasoning satisfies the two constraints. More details of SR$^3$ can be found in Algorithm~\ref{alg:algorithm_SR$^3$}.}
    \label{fig:reasoning_comparison}
\end{figure}

\section{Literature Review}

Large Language Models (LLMs), such as GPT-4o, offer comprehensive and distinct advantages for fake news detection~\cite{pan-etal-2023-fact,chen2024combating}. Benefiting from the vast amounts of real-world training data and advanced AI techniques (e.g., Transformer Framework, Reinforcement Learning), LLMs have demonstrated their powerful NLP abilities in perspectives of fact-checking~\cite{pan-etal-2023-risk}, text paraphrasing~\cite{qiang-etal-2023-parals, qiang2023chinese}, pattern recognition, and other CoT-recommendable tasks, such as question-answering~\cite{zhang2024tree}, explanation generation~\cite{bhattacharjee2024towards, wan-etal-2024-dell}, reasoning~\cite{jin2024cladder}, etc. One of the up-to-date studies utilizes GPT-3.5 to generate the reasoning with respect to news' common sense and description.~\cite{hu2309bad}. LLMs also can diversely solo fake news detection via few-shot~\cite{zeng2024justilm,ye2022unreliability} or zero-shot~\cite{mitchell2023detectgpt} prompting. A recent representative research~\cite{zhang2023towards} aims to mitigate the omission of thoughts and fact hallucination in fake news detection that merely manipulates LLMs for the news' claim verification by taking advantage of in-context learning (ICL), CoT, and external knowledge. In summary, LLMs have manifested their superiority in fake news detection via various prompting tips, yet their powerful usability in fake news detection is still being explored. 

Compared with the most recent related studies, this paper proposes a self-reinforced reasoning rectification approach (SR$^3$) that can iteratively rate the utility of generated positive and negative reasoning by feeding controversies into the next prompting iteration. Fig.~\ref{fig:reasoning_comparison} exhibits the process of our proposed LLM-based reasoning generation approach for news. It is observable that the SR$^3$ acts as a reasoning reviser by assessing the polarity (i.e., real or fake) and confidence level (i.e., how much real or fake) of reasoning-based news to verify if and how the generated reasoning can impact the original news. 

\section{Data Pre-Processing}

\subsection{Problem Formalization}
Suppose a randomly selected news and the corresponding label are denoted by $<x, y>$, where $y\in \{Real, Fake\}$. To ensure the effectiveness of generated reasoning in impacting the credibility of news, we first utilize a locally deployed LLM (OLlama 3 70B) to rate the credibility score of $x$, denoted as $V_{initial}$, where $V$ ranges from 0 (fake) to 100 (real). Then, we produce positive and negative reasoning by feeding LLM with multiple arguments in prompts. Eq.\ref{eq:initial_reasonings} illustrates the insights of the reasoning generation.

\begin{equation}
\begin{aligned}
    \Psi(x, y, T, V_{initial}, S) = (R, V), \\ 
    (R, V) \in \{(R^p, V^p),(R^n, V^n)\}
    \label{eq:initial_reasonings}
\end{aligned}
\end{equation}

where $\Psi$ embraces a prompt template that takes five arguments as inputs: (1) the news $x$, (2) corresponding label $y$, (3) the type of requested reasoning $T \in \{positive, negative\}$, (4) the credibility score of non-reasoning-based news $V_{initial}$, and (5) the alteration states $S \in \{increase, decrease\}$ for credibility score. Then, the manufactured $\Psi_{initial}$ is sent to LLM for reasoning requests. Finally, the LLM outputs: a pair of reasoning and evaluated credibility score -  $(R, V) \in \{(R^p, V^p),(R^n, V^n)\}$, where $(R^p, V^p)$ denote positive reasoning and corresponding credibility score, and vice versa. To make the generated reasoning available to use, we design a self-reinforced reasoning rectification (SR$^3$) method for valid reasoning generation. Algorithm~\ref{alg:algorithm_SR$^3$} describes the implementation details of SR$^3$.

\begin{algorithm}[ht]
\caption{Self-Reinforced Reasoning Rectification}
\label{alg:algorithm_SR$^3$}
\textbf{Input}: News item - $x$; Label - $y$;  the initial credibility score - $V_{initial}$; the type of requested reasoning $T$; the alteration states $S$ for credibility score; the pair of initial positive reasoning and credibility score - $(R^p, V^p)$; \\ the pair of initial negative reasoning and score ($R^n$, $V^n$) \\
\textbf{Parameter}: Polarity threshold of credibility score - $M$; \\
Expected incrementation for confidence level - $I$; \\
Maximum number of iterations - $Max\_Iter$\\
\textbf{Output}: The qualified reasoning $\{R^p, R^n\}$

\begin{algorithmic}[1] 

\IF {($y$ == '\textcolor{blue}{$fake$}' / '\textcolor{red}{$real$}')} 
\IF{$T$ == '$Positive$'}
\STATE $S \leftarrow $'\textcolor{blue}{$decrease$}' / ~'\textcolor{red}{$increase$}'; 
\WHILE{($V^p$ \textcolor{blue}{$>$} $M$) \& ($V^p - V_{initial}$ \textcolor{blue}{$>$} $I$ ) / '\textcolor{red}{$<$}'}
\STATE \textbf{Prompting} $\Psi_{re}(x, y, T, V_{initial}, S, R^p, V^p)$ ;
\STATE \textbf{\qquad return} $\{NewR^p, NewV^p\}$;
\STATE $R^p \leftarrow NewR^p$;
\STATE $V^p \leftarrow NewV^p$;
\STATE \textbf{UNTIL}  ~ $Max\_Iter$
\ENDWHILE
\ENDIF

\IF{T == 'Negative'}
\STATE $S \leftarrow $'\textcolor{blue}{$increase$}' / ~'\textcolor{red}{$decrease$}'; 
\WHILE{($V^p - V_{initial}$ \textcolor{blue}{$<$} $I$) /~'\textcolor{red}{$>$}'}
\STATE \textbf{Prompting} $\Psi_{re}(x, y, T, V_{initial}, S, R^n, V^n)$ ;
\STATE \textbf{\qquad return} $\{NewR^n, NewV^n\}$;
\STATE $R^n \leftarrow NewR^n$;
\STATE $V^n \leftarrow NewV^n$;
\STATE \textbf{UNTIL} ~ $Max\_Iter$
\ENDWHILE
\ENDIF
\ENDIF

\STATE \textbf{return} $\{R^p, R^n\}$
\end{algorithmic}
\end{algorithm}

Algorithm~\ref{alg:algorithm_SR$^3$} takes all the inputs and outputs of $\Psi$ as its inputs including $\{x, y, T, V_{initial}, S\}$, and $\{R^p, V^p, R^n, V^n\}$. The algorithm first checks the label of a given news, for example, if the label is fake (the conditions for fake news are in \textcolor{blue}{blue}, and that for real news are marked with \textcolor{red}{red} color in the algorithm). Then, we evaluate the initially generated positive reasoning $R^p$ (See lines 2-11). The alteration state S is confirmed as $'decrease'$ since positive reasoning can decrease the credibility score of fake news (See line 3). And, we set up two restrictions to inspect the quality of the generated positive reasoning in a 'While' loop, where the restrictions represent the polarity (symbolized by $M$) and confidence level (symbolized by $I$) of news, respectively (See line 4). Specifically, the polarity threshold $M$ (initially set to 50) is used to maintain the polarity of the label-specified news $x$, and the confidence level $I$ (initially set to 0) can restrict the credibility score of news being decreased while conducting positive reasoning for fake news. It is worth noting that the credibility score of fake news (or real news) is required to be decreased (or increased) as long as the expected positive reasoning $R^p$ is returned, and the decrementation of positive reasoning-based fake news (or incrementation of positive reasoning-based real news) is restricted by the confidence level $I$, vice versa for real news. Moreover, the procedure of generating fake news' negative reasoning (a.k.a., $y$ is 'fake' and $T$ is 'negative') is illustrated in lines 12-21. In this case, the alteration state S is transformed to $'increase'$ aiming to decrease the confidence level of the fake news by increasing its credibility score ($V^p$), yet we don't demand to alter the polarity of the negative reasoning-based news (line 14), vise versa for real news. Later, the re-generated $R^n$ will be sent back to LLMs via prompting for re-evaluating the credibility score of news until satisfying restrictions. Finally, the positive reasoning and negative reasoning that satisfy the two restrictions will be returned by the algorithm. 

\begin{figure*}
    \centering
    \includegraphics[width=1\linewidth,height = 0.51\linewidth]{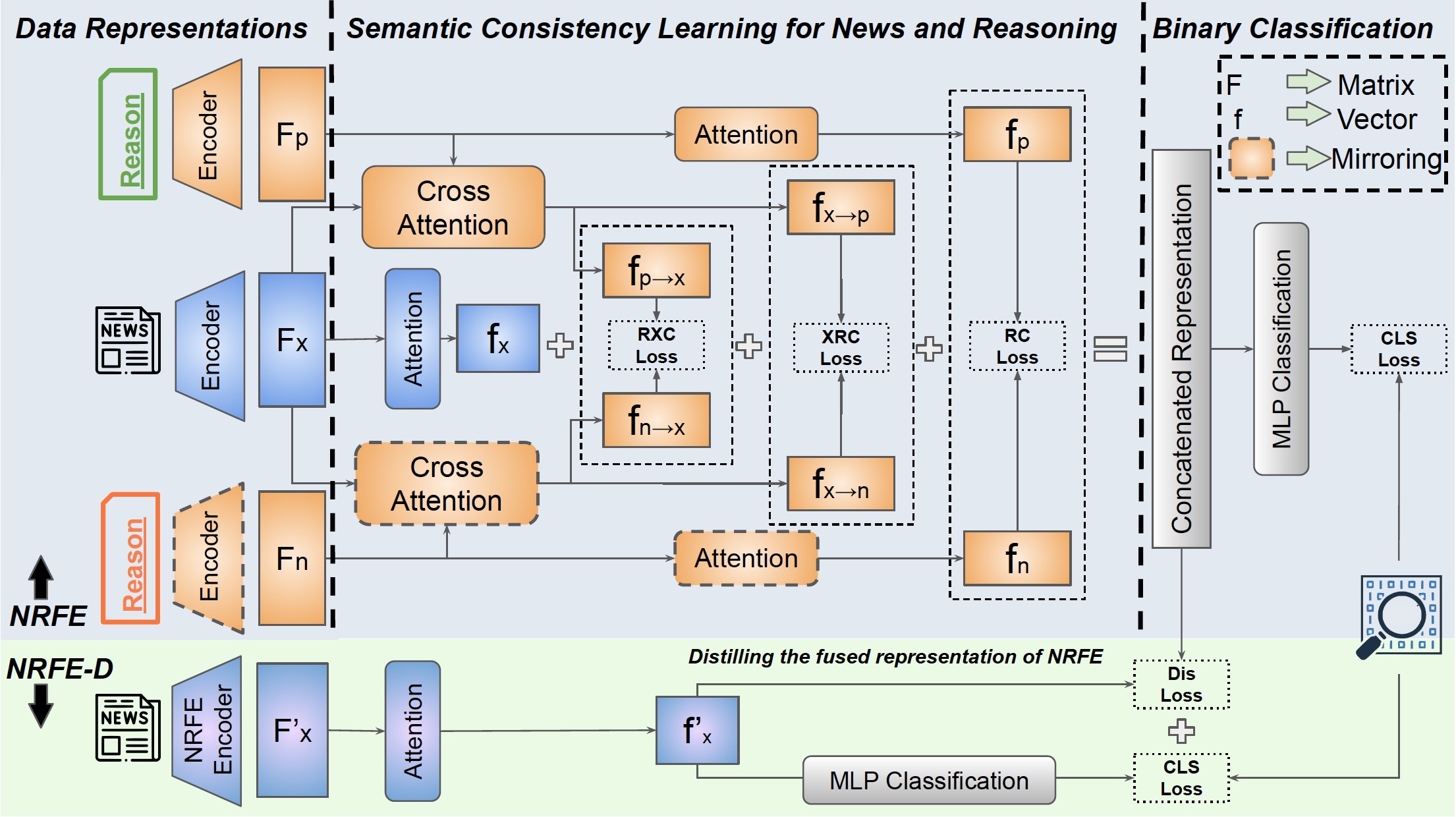}
    \caption{The system frameworks of \emph{NRFE} and \emph{NRFE-D}: the two Encoders embraced in \emph{NRFE} are used to represent news and reasoning, respectively. The dashed blocks in orange color are the mirroring of the blocks in orange color, which means they are parameter shared. For the knowledge distillation model - \emph{NRFE-D}, its Encoder used for news representation inherits the parameters of the news Encoder fine-tuned in \emph{NRFE}. \emph{NRFE-D} has two learning objectives: the fused representation from \emph{NRFE} (Dis Loss: $\mathcal L_{dis}$) and the hard label of data (CLS Loss: $\mathcal L'_{cls}$).}
    \label{fig:Framework}
\end{figure*}

\subsection{Datasets}
In this study, we deploy three widely adopted existing fake news datasets to conduct our experiment - Politifact~\cite{wang2017liar}, and Twitter-15 \& 16~\cite{yuan2019jointly}. More specifically, we leverage our proposed SR$^3$ to generate high-quality positive and negative reasoning for each news item from the three datasets. It is noteworthy that the labels of data items in two Twitter datasets are in four categories (i.e., non-rumors, true-rumors, false-rumors, and unverified-rumors). To adapt the settings of our method, we group the news items labeled by non-rumors and false-rumors as true samples, and the other two types of items are treated as false samples.

\section{Methodology}

\subsection{Data Representations}
Given a randomly selected news $x$ and the corresponding qualified positive \& negative reasoning $\{R^p, R^n\}$, it is capable of forming them as a positive news-reasoning pair - $(x, R^p)$ and a negative pair - $(x, R^n)$. \textit{NRFE} initially deploys two pre-trained BERT models as the Encoders for encoding news and reasoning, respectively. The Encoders can output sequential representations for $x$, and $R^p$ (or $R^n$), which are denoted as $F_x$, and $F_p$ (or $F_n$), respectively. 

To investigate the relationship between news and reasoning, we leverage a cross-attention block to facilitate their collaboration. The details of the interaction between news and reasoning are as shown in Eq.~\ref{eq:cross_attention}.

\begin{equation}
Cross Attention
    \begin{cases}
      (F_x, F_p) = \{ F_{p \rightarrow x}, F_{x \rightarrow p}\}\\
      (F_x, F_n) = \{F_{n \rightarrow x}, F_{x \rightarrow n}\}
    \end{cases} 
    \label{eq:cross_attention}
\end{equation}
where $F_{p \rightarrow x}$ and $F_{n \rightarrow x}$ represent the news article $x$ enriched with the information from positive reasoning $R^p$ and negative reasoning $R^n$, respectively. In contrast, $F_{x \rightarrow p}$ (or $F_{x \rightarrow n}$) captures the relevant information from the positive reasoning (or negative reasoning) based on the queries from the news. 

\subsection{Semantic Consistency Learning}

Since semantic gaps might exist between news and reasoning, it is necessary to transform the representations of news and reasoning into a common feature space for semantic regularization. To achieve this objective, we conduct semantic consistency learning to align the semantics between news and reasoning. Specifically, we present three binary classification tasks to predict whether news and reasoning have a semantic correlation or not. To simplify the procedure of semantic consistency learning task between news and reasoning, we fed the seven sequential representations - $F_x$, $F_p$, $F_n$, $F_{p \rightarrow x}$, $F_{n \rightarrow x}$, $F_{x \rightarrow p}$, and $F_{x \rightarrow n}$ to an Attention layer to acquire their global representations - $f_x, f_p, f_n, f_{p \rightarrow x}, f_{n \rightarrow x}, f_{x \rightarrow p}$, and $f_{x \rightarrow n}$. 

Given a news with corresponding positive and negative reasoning, we set the semantic correlation between news and positive reasoning as $1$, and the correlation between news and negative one as $0$. Then, we conduct three sub-tasks to align the semantic representations between news and reasoning - (1) the semantic consistency between news ($f_x$) and reasoning-aware news ($f_{p \rightarrow x}$ or $f_{n \rightarrow x}$); (2) the correlation between news ($f_x$) and news-aware reasoning ($f_{x \rightarrow p}$ or $f_{x \rightarrow n}$); (3) the correlation between news ($f_x$) and reasoning ($f_p$ or $f_n$). Each of the three sub-tasks is trained on an independent MLP-based classifier. In these three sub-tasks, we adopt cosine embedding loss with a preset hyperparameter - margin $d$ as the loss function (Eq.~\ref{eq:global_representations}) for training. 

\begin{equation}
\mathcal L_{rc} = 
\begin{cases}
     1-cos({f_x}, {f_r}), & if \; y = 1\\
     max(0, cos({f_x}, {f_r})-d), & if \; y = 0\\
\end{cases}  
\label{eq:global_representations}
\end{equation}
where $\mathcal L_{rc}$ is used to control the correlation between  $f_x$ and $f_r$, where $f_r \in \{f_p, f_n\}$. More specifically, $\mathcal L_{rc}$ can maximize the correlation between $f_x$ and $f_p$, and minimize the correlation between $f_x$ and $f_n$. 

Similarly, the correlation between news $r_x$ and reasoning-aware news $f_{r \rightarrow x} \in \{f_{p \rightarrow x}, f_{n \rightarrow x}\}$ is restricted by $\mathcal L_{rxc}$, and the correlation between news $f_x$ and news-aware reasoning $f_{x \rightarrow r} \in \{f_{x \rightarrow p}, f_{x \rightarrow n}\}$ is learned by $\mathcal L_{xrc}$. In this case, the representations of news, reasoning, and the interactions between them can be aligned into shared semantic spaces as the progress of training. Thus, the objective of semantic consistency learning $\mathcal L_{c}$ is as follows.

\begin{equation}
\begin{aligned}
\mathcal L_{c} =  \mathcal L_{rc} + 
                  \mathcal L_{rxc} + 
                 \mathcal L_{xrc}, \\
\end{aligned}
\end{equation}

Once the training phase of the semantic correlation module is completed, the positive pair ($\{f_x, f_p\}$) is again fed into the three well-trained blocks to obtain semantically aligned representations including news representation $m_x$, positive reasoning representation $m_p$, positive reasoning-aware news representation $m_{p \rightarrow x}$, and news-aware reasoning representation $m_{x \rightarrow p}$. Finally, these four representations will be composed as an entire representation (See Eq.\ref{eq:concatenation}), and the composition will be fed into the coming fusion module.

\begin{equation}
    m_{final} = m_x \oplus m_p \oplus m_{p \rightarrow x} \oplus m_{x \rightarrow p}
\label{eq:concatenation}
\end{equation}

Finally, \emph{NRFE} take the concatenation - $m_{final}$ as the input of an MLP to compare with the true label $y$, which can be formalized as follows.

\begin{equation}
\mathcal L_{cls} = CrossEntropy(y, MLP(m_{final})) \\
\label{eq:teacher_loss}
\end{equation}

\subsection{Binary Classification}
Since we adopt a supervised strategy to acquire reasoning from LLMs which means the prompt used for reasoning requests maintains news labels, it is unfair to directly feed news-reasoning pairs to \emph{NRFE} for model testing. Therefore, we design a reasoning-free model - \emph{NRFE-D} by acquiring the information from the hidden layers of \emph{NRFE} for knowledge distillation. Specifically, we first make \emph{NRFE-D} employing the news Encoder fine-tuned on \emph{NRFE} to accept news content as input for encoding the sequential representation of news $F^{'}_x$. Then, the vectorized global representation of news $f^{'}_x$ can be obtained via an Attention layer (similar to the transformation in \emph{NRFE}). And, $f^{'}_x$ goes through MLP to obtain the concatenated final representation $f^{'}_{final}$. Finally, we let $f^{'}_{final}$ approach to the final representation $m_{final}$ referenced from \emph{NRFE} by leveraging reverse KL loss~\cite{gu2024minillm}. Thus, the \emph{NRFE-D}'s learning objective from \emph{NRFE} is as follows (See Eq.~\ref{eq:reverse_KL}).

\begin{equation}
\begin{aligned}
       \mathcal L_{dis} &= KL(q \parallel p) \\
        &= \mathbb{E}_{x \sim p_x} \left[ \sum_{m_{final}} q(f^{'}_{final}|x) \log \frac{q(f^{'}_{final}|x)}{p(m_{final}|x)} \right] 
\label{eq:reverse_KL}
\end{aligned}
\end{equation}
where $p(m_{final}|x) = \frac{1}{N}\sum^{N}_{i=1}p(m^i_{final}|x^i)$ is the calculated probabilistic distribution from \emph{NRFE}, and $q(f^{'}_{final}|x)$ is the probabilistic distribution of \emph{NRFE-D}. To reconcile the impact of misclassification in \emph{NRFE}, we also leverage the 'hard' label of news (i.e., $y$) to train \emph{NRFE-D} using the cross-entropy loss (See Eq.~\ref{eq:CE_loss}).

\begin{equation}
\begin{aligned}
    \mathcal L'_{cls} = CrossEntropy(y, MLP(f^{'}_{final})) \\
\end{aligned} 
\label{eq:CE_loss}
\end{equation}
where the $f^{'}_{final}$ goes through a MLP to match the dimension of $y$ for training \emph{NRFE-D}. Thus, the total loss of \emph{NRFE-D} is:

\begin{equation}
    \mathcal L_D = \mathcal L_{dis} + \mathcal L'_{cls}
\end{equation}

\section{Experiments and Analysis}

\subsection{Baselines}
To highlight the superiority of our proposed approach, the baselines adopted in our experiments are three-folded, (1) prompting on LLMs (LLM-only), (2) fine-tuning on pre-trained SLMs (SLM-only), and (3) task-specific fake news detection methods. Since the label of news is used to request the reasoning, all of the baselines are verified using news content only for prediction as well as \emph{NRFE-D}.

\textbf{C1: LLM-only baselines} The LLMs we adopted to conduct experiments include OLlama 3 \textit{70B}, Gemma 2~\textit{27B}, and Mistral \textit{7B}, which are source-opened and locally deployable. The utilization of all the LLMs is under the setting of CoT and zero-shot. The performance of LLMs in fake news detection are evaluated across all the testing datasets.

\textbf{C2: SLM-only baselines} We deploy three Transformer Encoder-based bidirectional pre-trained language models - \textit{BERT}, \textit{RoBERTa}, and \textit{ALBERT} as baselines for performance comparison. It is worth noting that these SLMs are trained on the datasets via full-parameter fine-tuning. 

\textbf{C3: Task-specific baselines} 
\begin{itemize}
    \item \textbf{HSA} builds a hierarchical bidirectional LSTM to represent learning from the original posts and replies~\cite{guo2018rumor}. 
    
    \item \textbf{BiGRU} is one of the base models in the study~\cite{zhang2021mining}, which leverages BiGRU as the text Encoder to examine whether the features of publisher emotion and social emotion can facilitate the performance of fake news detection.
    
    \item \textbf{MUSER} is an evidence retrieval enhancement-based fake news detection method implemented by the sequence of summarization, retrieval, and reasoning~\cite{liao2023muser}. 
\end{itemize}

\subsection{Implementation Details}
In our experiments, we set up fixed hyper-parameters including the learning rate in Adam optimizer ($3^{-5\times 10}$), dropout rate ($0.3$), and the number of epochs ($30$ times) for both the baselines and our model. Each dataset is divided as 80\% for training and 20\% for testing. The metrics we adopted include accuracy (Acc.), macro F1 score (MacF1), precision (P-T for real news and P-F for fake news) as well as recall and F1 score, as shown in Tables \ref{tab:politi}-\ref{tab:Twitter-16}.

\subsection{Experimental Observations}
We have several observations from the results demonstrated in Tables~\ref{tab:politi}-\ref{tab:Twitter-16}.

\begin{table}[ht]\scriptsize
    \centering
    \begin{tabular}{cp{0.43cm}p{0.43cm}p{0.43cm}p{0.43cm}p{0.43cm}p{0.43cm}p{0.43cm}p{0.43cm}p{0.43cm}}
    \toprule
        \multirow{4}{*}{\bf \footnotesize Methods} & \multicolumn{8}{c}{\multirow{2}{*}{\bf \footnotesize PolitiFact Dataset}} \\ 
         &&&&&&& \\ \cline{2-9}
         & \multirow{2}{*}{Acc.}&\multirow{2}{*}{MacF1}&\multirow{2}{*}{P-T}&\multirow{2}{*}{R-T}&\multirow{2}{*}{F1-T}&\multirow{2}{*}{P-F}&\multirow{2}{*}{R-F}&\multirow{2}{*}{F1-F} \\
         & &&&&&&& \\ \cline{2-9}
          & &&&&&&& \\ 
        Llama 3  & 0.769 & 0.750 & 0.727 & 0.640 & 0.680 & 0.790 & 0.850 & 0.819 \\ 
        Gemma 2  & 0.622 & 0.509 & 0.518 & 0.186 & 0.274 & 0.639 & 0.892 & 0.744 \\ 
        Mistral  & 0.650 & 0.627 & 0.558 & 0.513 & 0.535 & 0.702 & 0.739 & 0.720 \\ 
        BERT     & 0.825 & 0.814 & 0.782 & 0.757 & 0.768 & 0.852 & 0.867 & 0.859\\ 
        RoBERTa  & 0.810 & 0.794 & 0.784 & 0.702 & 0.736 & 0.829 & 0.877 & 0.851 \\ 
        ALBERT   & 0.816 & 0.804 & 0.769 & 0.744 & 0.755 & 0.844 & 0.861 & 0.852 \\ 
        HSA & 0.785 & 0.764 & 0.766 & 0.637 & 0.693 & 0.796 & 0.877 & 0.834 \\ 
        BiGRU    & 0.740 & 0.714 & 0.702 & 0.592 & 0.631 & 0.770 & 0.833 & 0.797 \\ 
        MUSER    & 0.846 & 0.834 &\bf0.835& 0.746&0.788 &0.852 &\bf0.909 & 0.880 \\ 
        \bf \emph{NRFE-D} &\bf0.857&\bf0.847&0.831&\bf0.792&\bf0.808&\bf0.876&0.897&\bf0.885 \\ \bottomrule
    \end{tabular}
    \caption{Performance comparison between the baselines and \emph{NRFE-D} on the PolitiFact across multiple evaluation metrics. \emph{NRFE-D} achieves the highest performance in almost all metrics (e.g., highest Accuracy - 0.857 and Macro F1-score - 0.847), demonstrating its robust overall performance and superiority over other baselines. MUSER and ALBERT show strong performance, but they still lag behind NRFE-D.}
    \label{tab:politi}
\end{table}

First, The open-source LLMs (Llama 3, Gemma 2, Mistral) consistently generally yield the lowest results in terms of all the metrics across the three datasets compared with other methods. The phenomenon of their difficulty in adapting to fake news detection could be caused by their lack of specialization for the task-specific requirements and the potential inadequacy of their pre-training data for this specific application. More insightful, source-opened LLMs are versatile and general-purpose, but they lack task-specific training, affecting their performance in fake news detection.

\begin{table}[ht]\scriptsize
    \centering
    \begin{tabular}{cp{0.43cm}p{0.43cm}p{0.43cm}p{0.43cm}p{0.43cm}p{0.43cm}p{0.43cm}p{0.43cm}p{0.43cm}}
    \toprule
        \multirow{4}{*}{\bf \footnotesize Methods} & \multicolumn{8}{c}{\multirow{2}{*}{\bf \footnotesize Twitter-15 Dataset}} \\ 
         &&&&&&& \\ \cline{2-9}
         & \multirow{2}{*}{Acc.}&\multirow{2}{*}{MacF1}&\multirow{2}{*}{P-T}&\multirow{2}{*}{R-T}&\multirow{2}{*}{F1-T}&\multirow{2}{*}{P-F}&\multirow{2}{*}{R-F}&\multirow{2}{*}{F1-F} \\
         & &&&&&&& \\ \cline{2-9}
          & &&&&&&& \\ 
        Llama 3  & 0.709 & 0.708 & 0.727 & 0.657 & 0.690 & 0.695 & 0.760 & 0.726 \\ 
        Gemma 2  & 0.594 & 0.588 & 0.614 & 0.479 & 0.538 & 0.582 & 0.706 & 0.638 \\ 
        Mistral  & 0.641 & 0.637 & 0.672 & 0.534 & 0.595 & 0.622 & 0.746 & 0.678 \\ 
        BERT     & 0.918 & 0.918 & 0.906 & 0.931 & 0.918 & 0.931 & 0.906 & 0.918 \\ 
        RoBERTa  & 0.940 & 0.940 & 0.931 & 0.950 & 0.940 & 0.951 & 0.930 & 0.941 \\ 
        ALBERT   & 0.917 & 0.917 & 0.908 & 0.926 & 0.917 & 0.926 & 0.909 & 0.919 \\ 
        HSA & 0.822 & 0.821 & 0.803 & 0.849 & 0.824 & 0.847 & 0.794 & 0.818 \\ 
        BiGRU    & 0.889 & 0.889 & 0.863 & 0.920 & 0.891 & 0.917 & 0.858 & 0.887 \\ 
        MUSER   & 0.952 & 0.952 & 0.958 &\bf0.945& 0.951 &\bf0.947& 0.960 & 0.953 \\ 
        \bf \emph{NRFE-D} &\bf0.958&\bf0.959&\bf0.971&0.943&\bf0.957&\bf0.947&\bf0.972&\bf0.960\\ \bottomrule
    \end{tabular}
    \caption{Performance comparison on Twitter-15 dataset.}
    \label{tab:Twitter-15}
\end{table}

Second, fine-tuning on Pre-trained Language Models (BERT, RoBERTa, ALBERT) performs much better than the open-source LLMs. For example, BERT achieves high results in MacF1 (0.814) and Acc. (0.825) on PolitiFact, and RoBERTa also demonstrates its effectiveness and robustness on two Twitter datasets. Specifically, fine-tuning these models on fake news datasets allows them to learn task-specific nuances and improve their detection capabilities. 

\begin{table}[ht]\scriptsize
    \centering
    \begin{tabular}{cp{0.43cm}p{0.43cm}p{0.43cm}p{0.43cm}p{0.43cm}p{0.43cm}p{0.43cm}p{0.43cm}p{0.43cm}}
    \toprule
        \multirow{4}{*}{\bf \footnotesize Methods} & \multicolumn{8}{c}{\multirow{2}{*}{\bf \footnotesize Twitter-16 Dataset}} \\ 
         &&&&&&& \\ \cline{2-9}
         & \multirow{2}{*}{Acc.}&\multirow{2}{*}{MacF1}&\multirow{2}{*}{P-T}&\multirow{2}{*}{R-T}&\multirow{2}{*}{F1-T}&\multirow{2}{*}{P-F}&\multirow{2}{*}{R-F}&\multirow{2}{*}{F1-F} \\
         & &&&&&&& \\ \cline{2-9}
          & &&&&&&& \\ 
        Llama 3  & 0.771 & 0.769 & 0.853 & 0.729 & 0.786 & 0.690 & 0.828 & 0.753 \\ 
        Gemma 2  & 0.542 & 0.536 & 0.692 & 0.375 & 0.486 & 0.473 & 0.771 & 0.586 \\ 
        Mistral  & 0.650 & 0.649 & 0.743 & 0.604 & 0.666 & 0.568 & 0.714 & 0.632 \\ 
        BERT     & 0.920 & 0.914 & 0.955 & 0.895 & 0.924 & 0.868 & 0.942 & 0.904 \\ 
        RoBERTa  & 0.934 & 0.933 & 0.949 & 0.937 & 0.943 & 0.916 & 0.931 & 0.923 \\ 
        ALBERT   & 0.906 & 0.903 & 0.917 & 0.920 & 0.918 & 0.890 & 0.885 & 0.888 \\ 
        HSA & 0.903 & 0.900 & 0.903 & 0.933 & 0.918 & 0.905 & 0.862 & 0.882 \\
        BiGRU    & 0.927 & 0.926 & 0.952 & 0.920 & 0.936 & 0.896 & 0.937 & 0.916 \\ 
        MUSER   & 0.938 & 0.939 & 0.957 & 0.937 & 0.947 & 0.916 &0.942 & 0.929 \\ 
        \bf \emph{NRFE-D} &\bf0.971&\bf0.970&\bf0.991&\bf0.958&\bf0.974&\bf0.945&\bf0.988&\bf0.966 \\ 
         \bottomrule
    \end{tabular}
    \caption{Performance comparison on Twitter-16 dataset.}
    \label{tab:Twitter-16}
\end{table}

Third, the task-specific models - HSA, BiGRU, and MUSER are designed to tackle the unique challenges of fake news detection by leveraging specific aspects like social information, emotional features, and evidence retrieval. Because our datasets lack social information and emotional features, HSA and BiGRU showcase their lower performance than fine-tuning on Pre-trained Language Models, yet demonstrate their strengths compared with LLMs. MUSER demonstrates its high performance across all of the metrics (0.846, 0.952, and 0.938 in Accuracy, and 0.834, 0.952, and 0.939 in MacF1 on the three datasets), making it particularly strong in detecting misinformation.

Finally, \emph{NRFE-D} achieves the relatively highest performance across the datasets compared with the other nine baselines (outperforms at least 1.1\% Accuracy and 1.3\% MacF1 on PolitiFact,  0.6\% Accuracy and 0.7\% MacF1 on Twitter15, and  3.3\% Accuracy and 3.1\% MacF1 on PolitiFact). By taking advantage of generated positive and negative reasoning via SR$^3$,  \emph{NRFE} leverages dual BERT Encoders to learn semantic consistency between news and reasoning, thereby enriching \emph{NRFE}'s logical and semantic comprehension for news. Based on that, \emph{NRFE-D} not only distills the knowledge from \emph{NRFE} but also doesn't abandon utilizing the original label of news, leading to consistent high performance across datasets and demonstrating its robustness and adaptability to diverse misinformation contexts.

\begin{figure*}[ht]
\centering
\subfigure[PolitiFact]{
\includegraphics[width=0.67\columnwidth]{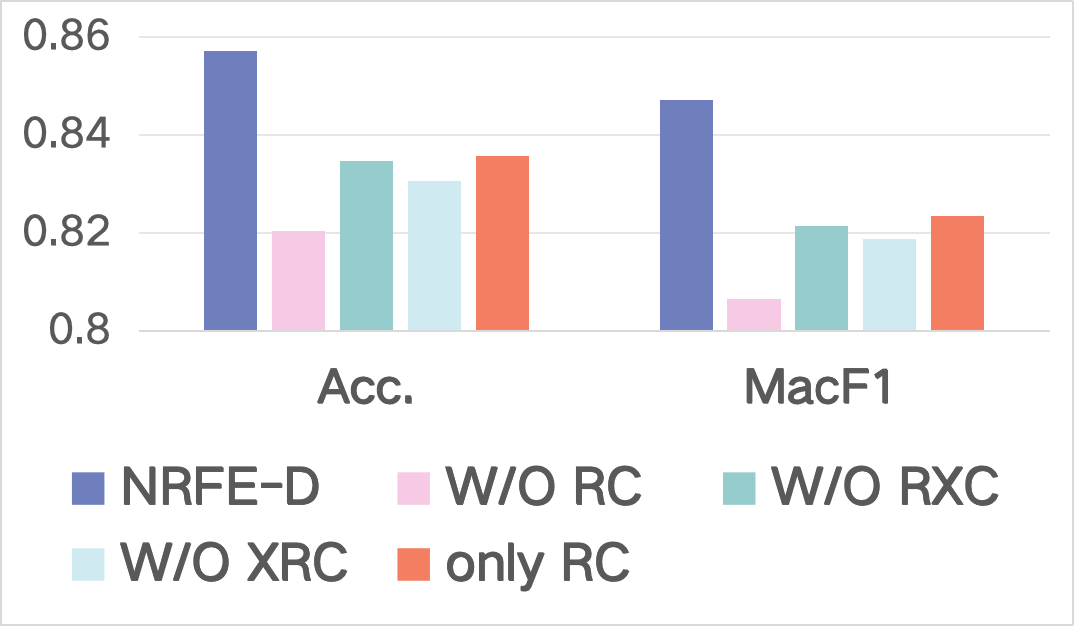}\label{fig:ablation_politi}}
\subfigure[Twitter-15]{ 
\includegraphics[width=0.67\columnwidth]{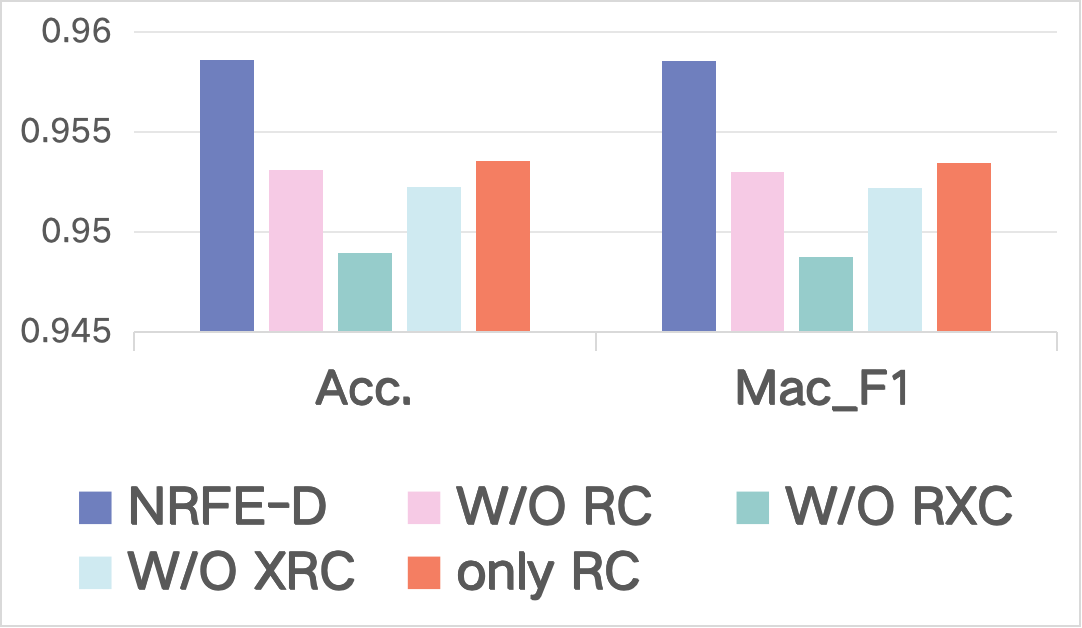}\label{fig:ablation_twitter15}}
\subfigure[Twitter-16]{ 
\includegraphics[width=0.66\columnwidth]{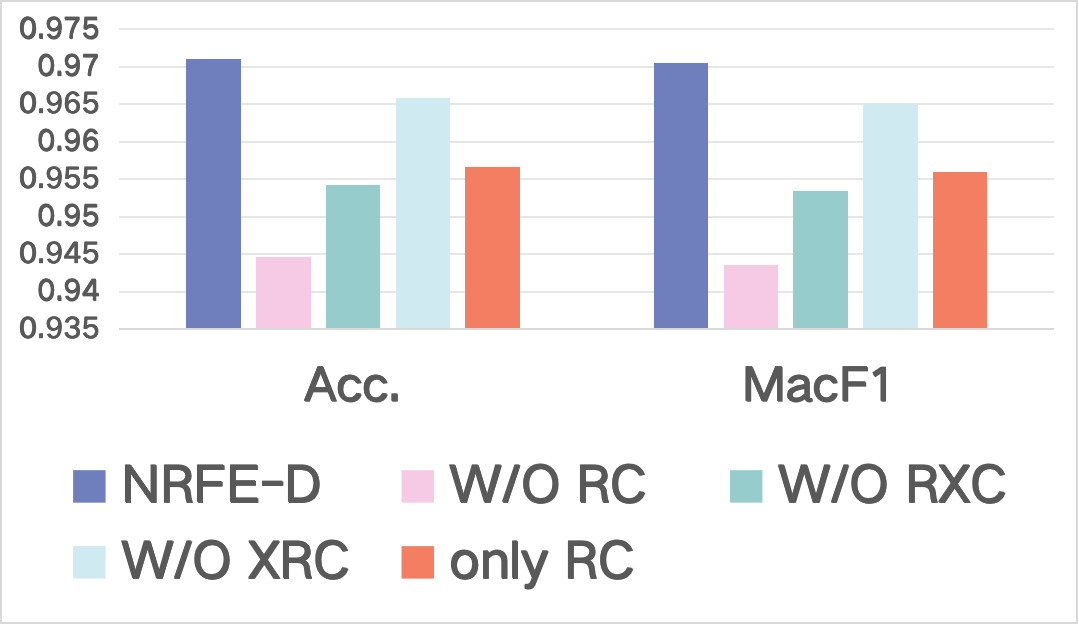}\label{fig:ablation_twitter16}}
\caption{Ablation comparisons among \emph{NRFE-D} and its variants in terms of accuracy and MacF1 on three datasets. RC is the RC loss used for learning the semantic correlation between news $f_x$ and reasoning $f_p$ or $f_n$ as well as RXC and XRC in Fig.~\ref{fig:Framework}. The last variant - Only RC, denotes that RXC and XRC are disabled for ablation evaluation.} 
\label{fig:ablation_study1}
\end{figure*}

\subsection{Ablation Study}
\begin{figure}[ht]
\centering
\subfigure[NRFE]{
\includegraphics[width=0.31\columnwidth]{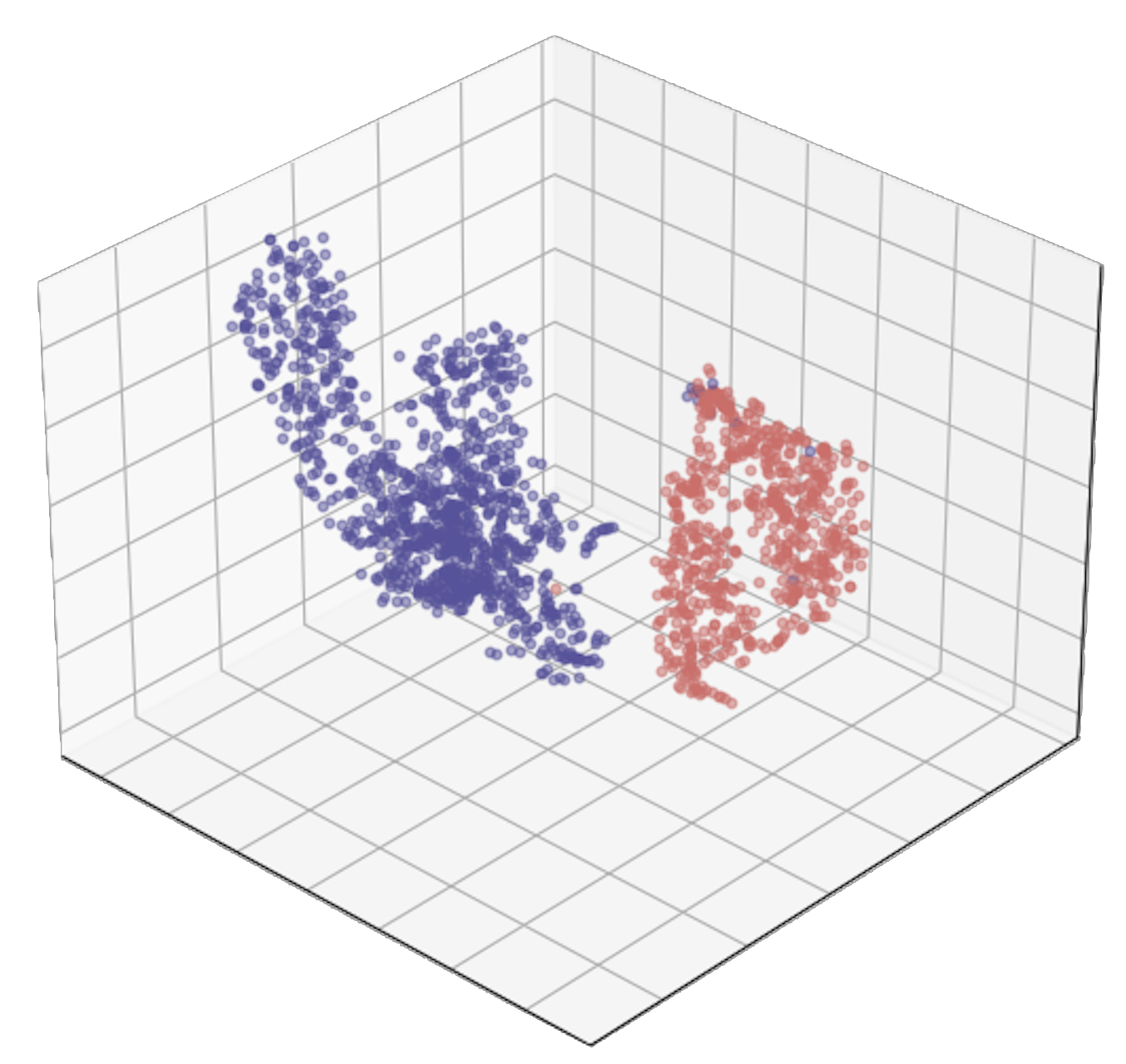}\label{fig:entire}}
\subfigure[w/o RC]{ 
\includegraphics[width=0.31\columnwidth]{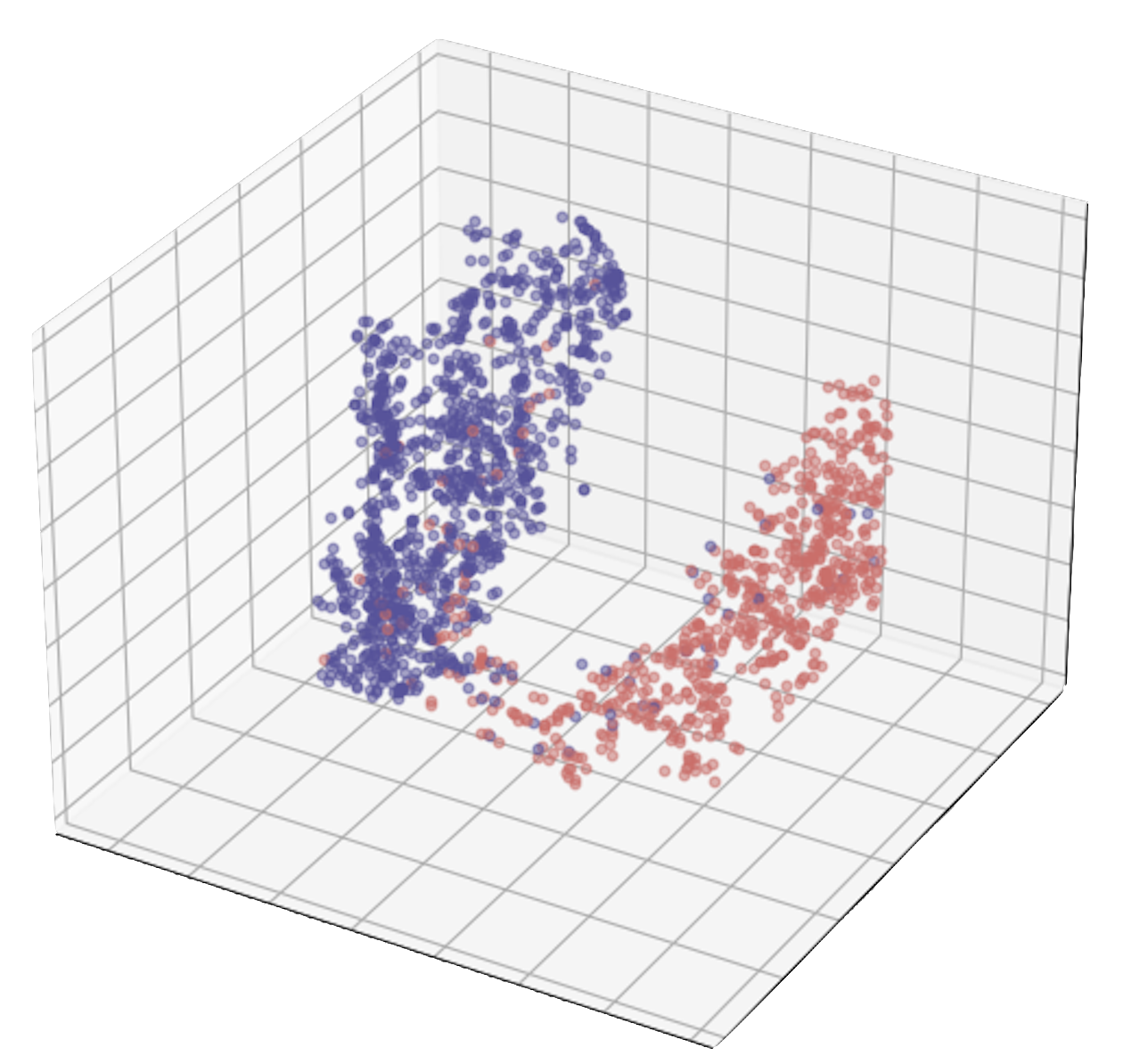}\label{fig:rc}}
\subfigure[w/o RXC]{ 
\includegraphics[width=0.31\columnwidth]{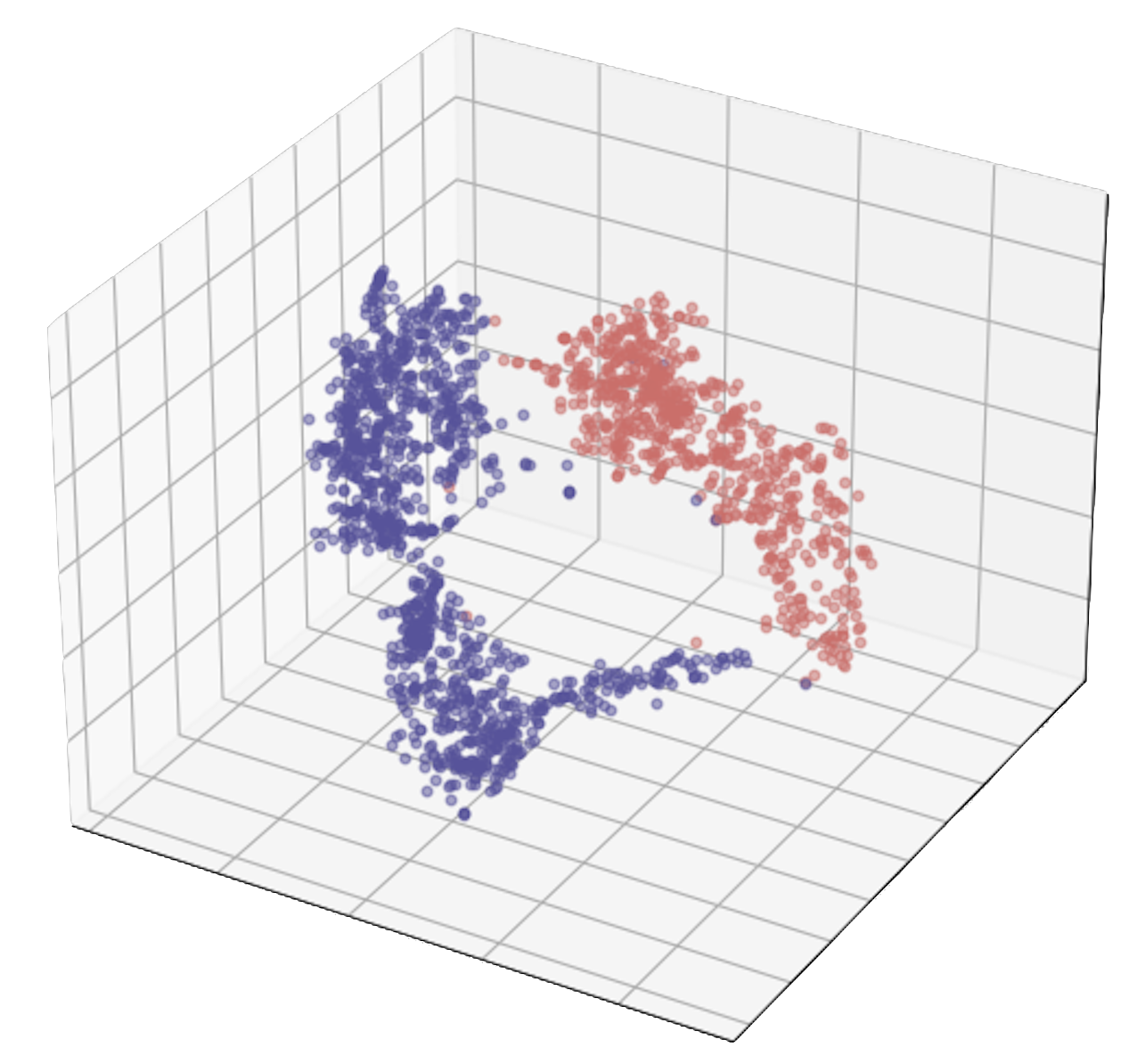}\label{fig:rxc}}
\subfigure[w/o XRC]{ 
\includegraphics[width=0.31\columnwidth]{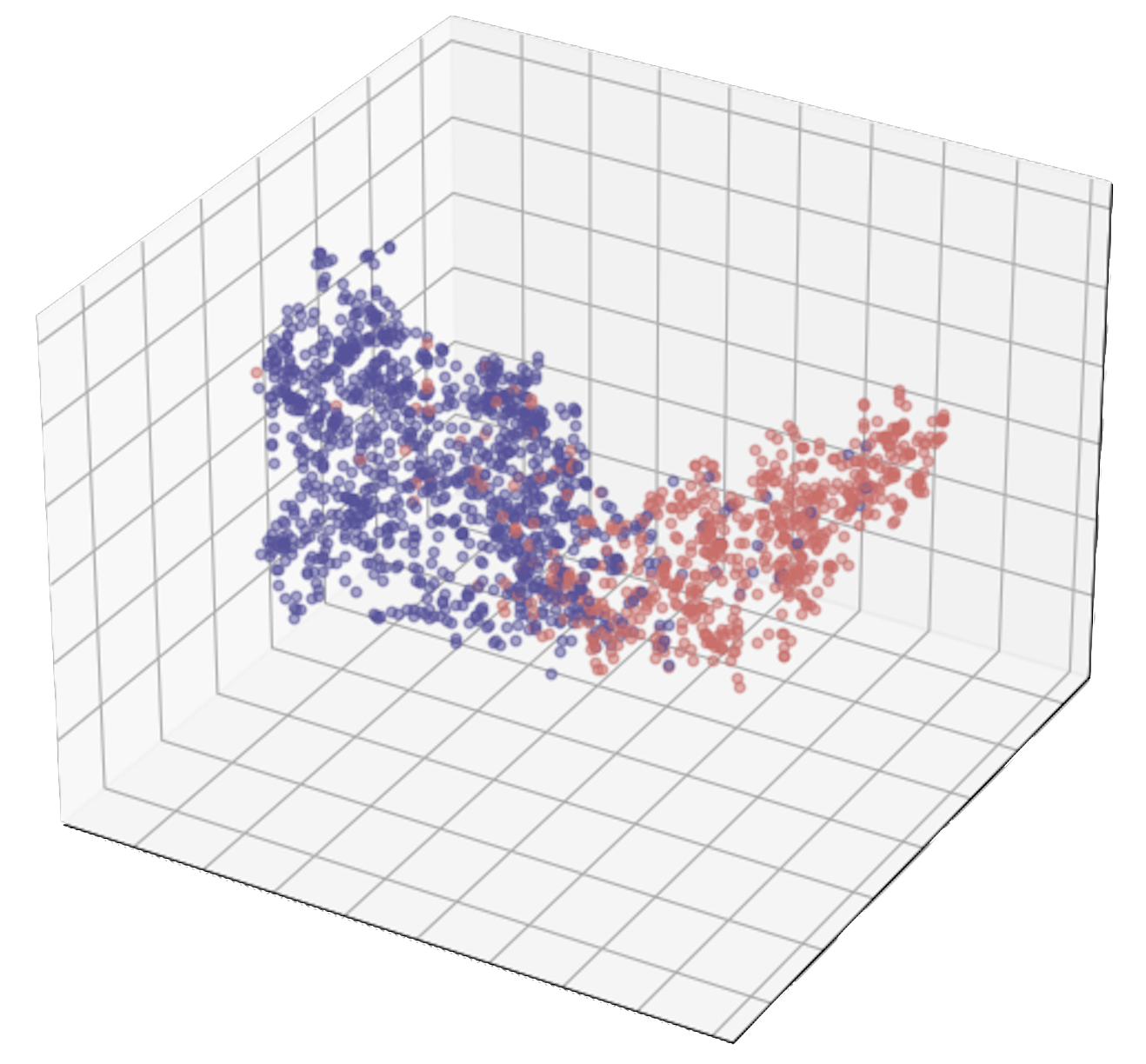}\label{fig:xrc}}
\subfigure[only RC]{ 
\includegraphics[width=0.31\columnwidth]{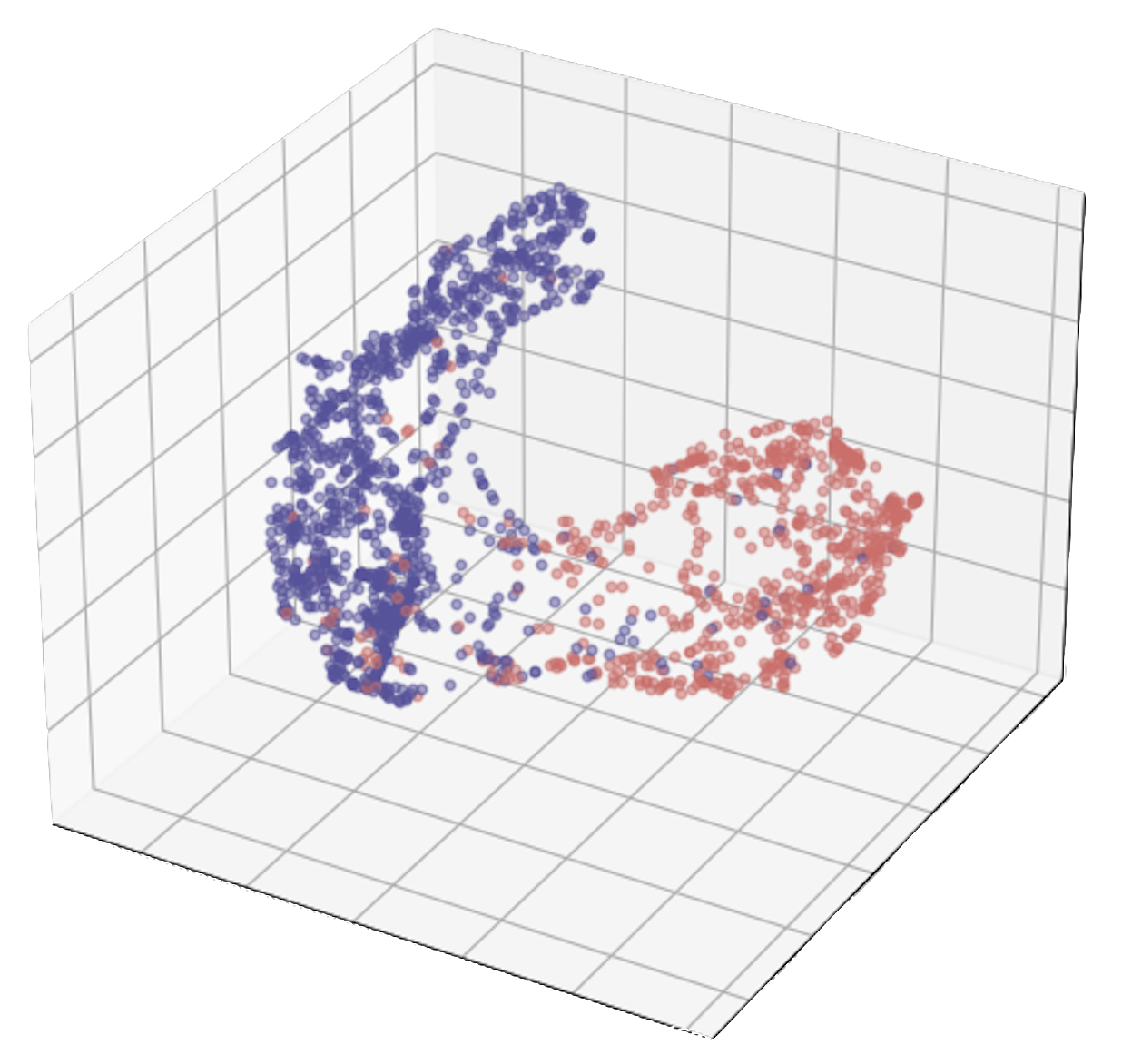}\label{fig:only_c}}
\caption{Ablation comparisons on PolitiFact dataset among the teacher model - \emph{NRFE} and its four variants by visualizing their features of the last hidden layer using T-SNE. The dots in blue color denote real news, and vice versa.} 
\label{fig:ablation_study2}
\end{figure}

We conduct ablation experiments on both \emph{NRFE} and \emph{NRFE-D}. Specifically, we evaluate the effectiveness of the different components in our model by disabling  RXC loss (i.e., w/o RXC), XRC loss (i.e., w/o XRC), RC loss (i.e., w/o RC), or RXC\&XRC loss (i.e., only RC) as shown in Figs.~\ref{fig:ablation_study1} and \ref{fig:ablation_study2}.

In Fig.~\ref{fig:ablation_study1}, all the different variants of \emph{NRFE-D} yield worse performance in terms of accuracy and MacF1 compared with \emph{NRFE-D}. We inspect the effectiveness of each block with the following observations. First, \emph{NRFE-D} w/o RC yields the relatively lowest performance on PolitiFact and Twitter-16 datasets, which demonstrates the importance of the semantic consistency between news and corresponding reasoning. In other words, \emph{NRFE-D} ensures the semantic alignment between them by taking advantage of RC, thereby reinforcing the understanding and improving the detection of fake news. Second, w/o RXC and w/o XRC shows a bit of improvement in two metrics compared with w/o RC on PolitiFact and Twitter-16. Yet, they show a noticeable performance decline compared with \emph{NRFE-D}. RXC and XRC can capture the interactions between news and reasoning under the cross-attention mechanism, which indicates their significance and contributions to the performance of fake news detection. Finally, only RC produces better performance than the other three variants on PolitiFact and Twitter-15, indicating the importance of RXC and XRC in \emph{NRFE-D}. 

Fig.~\ref{fig:ablation_study2} discover the role of different blocks in the training phase of the teacher model (i.e., \emph{NRFE}). It is observable that the margin between fake and real news in \emph{NRFE} is much broader than the other variants coupled with its better correctness ratio in classification, which means the features of training data learned by \emph{NRFE} are more discriminative than the others (See. Fig.~\ref{fig:entire}). The variant w/o RC demonstrates a significant gap between most positive and negative data samples, yet the rate of misclassification is more considerable than \emph{NRFE} (See. Fig.~\ref{fig:rc}). The layout of w/o RXC is more likely a triple-classification although it has significant correctness of training classification (See. Fig.~\ref{fig:rxc}). The variant - w/o XRC generates the worst results, which indicates the importance of news-related reasoning in affecting the performance of classification (See. Fig.~\ref{fig:xrc}). Overall, we can see from the figures that the aggregation of features from different blocks can significantly improve the representation ability for news data, thereby enhancing the effectiveness of model training.

\section{Conclusion}
Unlike existing LLM reasoning-based fake news detection approaches, we propose a novel self-reinforced reasoning rectification method (SR$^3$) to conduct the production of reasoning in a supervised manner, which leverages the ability of LLMs hallucination for negative reasoning generation. In detail, we drive OLlama 3 70B to generate high-quality positive and negative reasoning for news using SR$^3$ via LLM reflection. Upon that, we deploy a local Negative Reasoning-based FakE news detection model - \emph{NRFE} to learn the semantic consistency between news and reasoning (positive or negative). To evaluate the performance of our model, we present a student model - \emph{NRFE-D} by distilling the knowledge of \emph{NRFE} learned from news and reasoning, thereby avoiding the impact of surprised reasoning requests. The experiment results show that our method yields outstanding performance compared with different types of SOTA baselines. In the future, we will further investigate the potentiality of other source-opened LLMs (such as OLlama 3.1, or Mistral Large 2) in generating negative reasoning in an unsupervised manner. On the other hand, LLM-based multi-agents can also be transferred in the scenes of fake news detection, which will be one of our future study directions. 

\section{Acknowledgments}

We greatly appreciate the anonymous reviewers who provide insightful comments for this work. This study is partially supported by the National Natural Science Foundation of China (62403412, 62076217), the Natural Science Foundation of the Higher Education Institutions of Jiangsu Province of China under grant 23KJB520040, and the National Language Commission of China (ZDI145-71).

\bibliography{aaai25}

\end{document}